\documentclass{pibench}
\usepackage[utf8]{inputenc}
\usepackage{amsmath}
\usepackage{amssymb}
\usepackage{array}
\usepackage{booktabs}
\usepackage{colortbl}
\usepackage{float}
\usepackage{graphicx}
\usepackage{multirow}
\usepackage{tabularx}
\usepackage{url}
\usepackage{xspace}
\usepackage{xcolor}
\usepackage{adjustbox}

\usepackage{amsmath,amsfonts,bm}

\def\eqref#1{equation~\ref{#1}}

\def\1{\bm{1}}

\DeclareMathAlphabet{\mathsfit}{\encodingdefault}{\sfdefault}{m}{sl}
\SetMathAlphabet{\mathsfit}{bold}{\encodingdefault}{\sfdefault}{bx}{n}

\title{Alipay-PIBench: A Realistic Payment Integration Benchmark for Coding Agents}

\author[1]{Shiyu Ying}
\author[1]{Xuejie Cao}
\author[1]{Yingfan Ma}
\author[1]{Yuanhao Dong}
\author[1]{Wenyu Chen}
\author[1]{Bowen Song}
\author[1]{Lin Zhu}
\affiliation[1]{Ant Group}
\correspondence{\email{lin.zhulin@antgroup.com}}
\gtechdata[Code]{\url{https://github.com/inclusionAI/PIBench}}

\newcommand{\bench}{Alipay-PIBench\xspace}
\newcommand{\withskill}{with-skill\xspace}
\newcommand{\noskill}{without-skill\xspace}
\newcommand{\dangmianfu}{QR Code Payment\xspace}

\newcommand{\appzhifu}{In-App Payment\xspace}

\newcommand{\yushouquanzhifu}{Authorization Hold \xspace}

\newcommand{\payskill}{\texttt{alipay-payment-integration}\xspace}

\newcommand{\claudeopus}{Claude Opus 4.8\xspace}
\newcommand{\glm}{GLM-5.2\xspace}
\newcommand{\kimicode}{Kimi K2.7 Code\xspace}
\newcommand{\deepseekpro}{DeepSeek-V4-Pro\xspace}
\newcommand{\minimax}{MiniMax M3\xspace}
\newcommand{\qwenthirlyseven}{Qwen3.7-Max\xspace}
\definecolor{deltaposlowbg}{RGB}{226, 240, 255}
\definecolor{deltaposmidbg}{RGB}{185, 217, 255}
\definecolor{deltaposhighbg}{RGB}{125, 183, 240}
\definecolor{deltaneglowbg}{RGB}{255, 240, 221}
\definecolor{deltanegmidbg}{RGB}{255, 210, 168}
\definecolor{deltaneghighbg}{RGB}{246, 162, 96}
\newcommand{\deltaposlow}[1]{\cellcolor{deltaposlowbg}#1}
\newcommand{\deltaposmid}[1]{\cellcolor{deltaposmidbg}#1}
\newcommand{\deltaposhigh}[1]{\cellcolor{deltaposhighbg}#1}
\newcommand{\deltaneglow}[1]{\cellcolor{deltaneglowbg}#1}
\newcommand{\deltanegmid}[1]{\cellcolor{deltanegmidbg}#1}

\newcommand{\deltaavg}[1]{\textbf{#1}}
\newcommand{\outputrellow}[1]{\cellcolor{deltaposlowbg}#1}
\newcommand{\outputrelmid}[1]{\cellcolor{deltaposmidbg}#1}
\newcommand{\outputrelhigh}[1]{\cellcolor{deltaposhighbg}#1}
\newcommand{\outputbadlow}[1]{\cellcolor{deltaneglowbg}#1}
\newcommand{\outputbadmid}[1]{\cellcolor{deltanegmidbg}#1}
\newcommand{\outputbadhigh}[1]{\cellcolor{deltaneghighbg}#1}
\newcommand{\rqhead}[1]{\begin{tabular}[c]{@{}c@{}}#1\end{tabular}}

\abstract{
Coding agents are increasingly being evaluated on realistic software development tasks that require repository-level editing, cross-component integration, domain knowledge, and task-specific verification. Payment integration is a representative domain-specific setting because a correct implementation must select an appropriate payment product, keep credentials and signing logic within server-side trust boundaries, coordinate frontend and backend flows, verify payment outcomes, and maintain consistency between payment and local business states.

We introduce \bench, a repository-level benchmark proposed by Alipay for evaluating AI coding agents on realistic payment integration tasks. Built around Alipay Open Platform products and business-oriented repositories, \bench evaluates functional correctness, reliability, security, and business-state consistency. The benchmark comprises nine product-specific projects and 18 task instances organized into two progressive scenarios: Basic (Functional Payment Completion), which evaluates the core end-to-end payment loop, and Advanced (Risk-Aware Payment Hardening), which evaluates agents' ability to harden payment integrations against payment-specific risks through notification idempotency, abnormal-transaction handling, refund safeguards, and fund-safety controls. Generated solutions are evaluated using deterministic checks derived from scenario-specific rubrics, supplemented by rubric-aligned LLM-assisted assessment. Across the six evaluated models, mean RPR under the \withskill condition ranges from 68.58\% to 91.37\%. Compared with the corresponding \noskill condition, the \withskill condition improves mean RPR by 10.31 percentage points on average, with gains varying across models, products, and scenarios. Method-level results further distinguish source-level completion, executable payment behavior, and payment-domain requirements.
}

\begin{document}
\maketitle

\section{Introduction}

Recent evaluations of coding agents are moving beyond isolated code generation toward more realistic software delivery tasks. Benchmarks such as HumanEval, SWE-bench, and LiveCodeBench have been effective for measuring function-level programming, general bug fixing, and repository-level code modification \citep{chen2021evaluating,jimenez2024swebench,jain2024livecodebench}. In real development workflows, however, many tasks require more than locally correct code: an agent must understand the business scenario, choose an appropriate domain solution, coordinate changes across components, and preserve safety and state consistency after implementation.

Payment integration is a typical example of this shift. For many applications, payment capability is a key step from product workflow to commercial operation. Yet integrating a payment product is not simply adding an API call. A correct implementation must select a suitable payment product, keep credentials and signing logic on the server side, coordinate frontend and backend payment flows, verify asynchronous notifications, and update local order or service state only after reliable payment confirmation. If an agent only makes the code appear runnable while trusting client-side payment status, skipping signature verification, mishandling repeated notifications, or causing inconsistency between payment state and business fulfillment, the result may introduce real fund-safety and business-consistency risks.

These properties make payment integration a challenging domain-specific setting for evaluating coding agents. It jointly tests code generation, cross-component integration, payment product understanding, risk handling, and business-state maintenance. Existing general-purpose benchmarks are not designed to systematically measure these capabilities, because the correctness of payment integration depends not only on local code behavior, but also on payment-domain rules, security boundaries, and consistency across the business workflow.

To address this setting, we propose \bench, an Alipay-led repository-level benchmark for evaluating AI coding agents on realistic Alipay payment integration tasks. Built on Alipay Open Platform products and business-style repositories, \bench evaluates not only whether an agent can complete a payment integration, but also whether the implementation is secure, reliable, convenient to integrate, and consistent with local business state.

\bench follows a systematic evaluation pipeline. First, we use \textbf{Product-Grounded Task Construction} to connect real payment products, business scenarios, initial repositories, and natural-language integration requests, so that each task evaluates both product understanding and repository-level implementation. Second, we divide payment integration into two progressive scenarios: \textbf{Basic (Functional Payment Completion)}, which focuses on the core payment loop including payment creation, payment interaction, backend confirmation, and local state update; and \textbf{Advanced (Risk-Aware Payment Hardening)}, which focuses on risk hardening such as signature verification, idempotency, abnormal transaction states, refunds, and fund-safety boundaries. Third, we adopt \textbf{Rubric-Derived Evaluation}, deriving deterministic checks and LLM-assisted assessment criteria from scenario-specific rubrics so that evaluation items remain traceable to payment-domain requirements rather than ad hoc tests or generic judge prompts. Finally, through a paired \textbf{\noskill/\withskill} setting, \bench evaluates whether the \payskill skill helps agents complete payment integration more efficiently, conveniently, and safely.

This paper makes the following contributions:
\begin{itemize}
    \item We propose the first Alipay-led repository-level benchmark for evaluating coding agents on realistic Alipay payment integration tasks.

    \item We propose \textbf{Product-Grounded Task Construction}, which organizes payment products, business scenarios, initial repositories, and natural-language integration requests into a unified task format.

    \item We introduce a progressive capability decomposition of payment integration into \textbf{Basic (Functional Payment Completion)} and \textbf{Advanced (Risk-Aware Payment Hardening)}, distinguishing end-to-end payment completion from the ability to harden payment integrations against payment-specific risks and boundary failures.
    
    \item We develop \textbf{Rubric-Derived Evaluation}, which derives deterministic checks and supplementary large-language-model (LLM)-assisted rubric assessment criteria from scenario-specific payment rubrics, making evaluation evidence traceable to domain requirements.

    \item We establish a paired \textbf{\noskill/\withskill} intervention protocol to measure how access to structured Alipay payment guidance changes adherence to product-specific APIs and integration patterns, payment safety, payment--business state consistency, and output-token efficiency.
\end{itemize}

Across the 18 task instances and six evaluated models, mean RPR under the \withskill condition ranged from 68.58\% to 91.37\%. Compared with the corresponding \noskill condition, the \withskill condition improved mean RPR by 10.31 percentage points on average, with gains in 101 of 108 model--product--scenario comparisons. The average improvement was larger for Basic tasks than for Advanced tasks (+11.27 versus +9.35 percentage points). Method-level results further distinguish source-level completion, executable payment behavior, and payment-domain requirements, motivating the use of progressive scenarios and complementary rubric-grounded evaluation signals.

\section{Related Work}

\paragraph{Repository-level benchmarks for coding agents.}
Coding benchmarks have expanded from function-level generation toward realistic software engineering. HumanEval, APPS, and LiveCodeBench evaluate isolated programming and algorithmic problem solving \citep{chen2021evaluating,hendrycks2021measuring,jain2024livecodebench}, while SWE-bench evaluates issue resolution in existing repositories through executable tests \citep{jimenez2024swebench}. Recent benchmarks further cover repository-level feature implementation, project construction, and longer-horizon development. FEA-Bench studies feature addition \citep{li2025feabench}; ProjDevBench and NL2Repo-Bench evaluate project or repository construction from natural-language requirements \citep{lu2026projdevbench,ding2026nl2repo}; and BeyondSWE broadens evaluation to cross-repository reasoning, domain-specific repair, dependency migration, and repository generation \citep{chen2026beyondswe}. DeepSWE further introduces original repository-scale tasks and behavior-oriented verifiers to reduce contamination and implementation-specific grading \citep{huang2026deepswe}. Domain-specific benchmark design is also exemplified by ABench-Physics, which evaluates model performance on high-difficulty, dynamic physics problems \citep{zhang2025abenchphysics}. These benchmarks reflect a transition toward realistic software delivery, but mostly remain organized around general software-engineering task forms.

\paragraph{Production and payment-integration evaluation.}
Production-oriented evaluations increasingly emphasize realistic repositories, deployment constraints, and reliable verification. ProdCodeBench derives executable tasks from production developer-agent sessions and applies test-relevance and stability checks during benchmark construction \citep{jha2026prodcodebench}. Security-oriented evaluations further examine whether functionally successful implementations satisfy deployment-critical security requirements \citep{endor2026agentsecurity}. Most closely related to the payment setting, Stripe reports an industry evaluation suite for coding agents on realistic Stripe integrations, using full coding environments and deterministic graders over APIs, user interfaces, and Stripe-side artifacts \citep{stripe2026integrationbenchmark}. Together, these efforts demonstrate a broader shift from generic code correctness toward production-grounded evaluation under domain and platform constraints. \bench follows this direction by studying repository-level payment integration in the Alipay ecosystem, with additional attention to payment-specific risks and business-state consistency.

\paragraph{Rubric-grounded evaluation and domain guidance.}
Executable tests provide reproducible evidence of software behavior, but may not fully capture product selection, security placement, or cross-component business consistency. Some recent benchmarks therefore combine execution-based evaluation with LLM-assisted review, although RuVerBench shows that LLM rubric verification in agentic coding remains noisy \citep{lu2026projdevbench,peng2026ruverbench}. In \bench, scenario-specific payment rubrics provide a common specification from which deterministic checks and supplementary semantic criteria are derived, while executable evaluation remains the primary correctness signal. Separately, Stripe's agent-steering experiments indicate that explicitly loaded, structured skills can guide integration behavior more effectively than passive warnings or documentation \citep{beswick2026steering}. Motivated by this practical question, \bench performs paired \noskill/\withskill trials under the same task, environment, and evaluation criteria to measure the effect of the Alipay payment skill.

\section{\bench Overview}

\begin{figure*}[t]
    \centering
    \includegraphics[width=\textwidth]{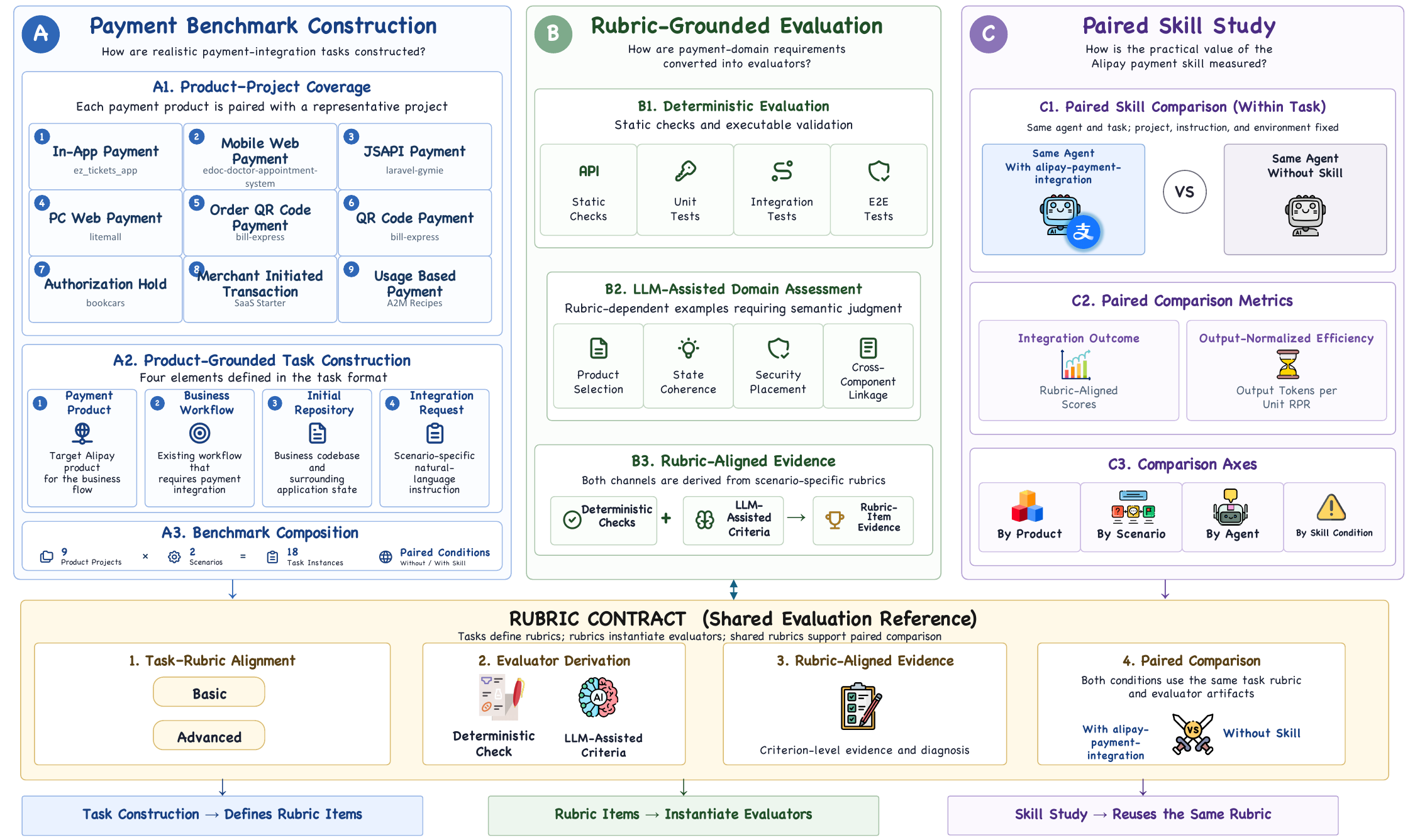}
    \caption{Overview of the Alipay-PIBench framework. Product-specific projects are paired with business workflows, initial repositories, and integration requests to form Basic and Advanced task scenarios. Deterministic checks and LLM-assisted criteria are derived from shared rubrics, while paired without-skill and with-skill conditions measure the practical effect of \payskill.}
    \label{fig:benchmark-framework}
\end{figure*}

\subsection{Benchmark Composition}

\bench is a benchmark suite for repository-level payment integration. It is organized around product-specific projects, each corresponding to one Alipay payment product and an associated business setting. The selected products span multiple common Alipay integration patterns, including mobile-app payments, web-based payments, in-person code payments, pre-authorization, and recurring or subscription-style payments. These products differ in user entry points, payment-initiation flows, backend confirmation paths, and risk profiles, allowing the benchmark to evaluate both product understanding and repository-level implementation.

Each product-specific project contains two progressive scenarios. \textbf{Basic (Functional Payment Completion)} starts from a business workflow in which the target Alipay product has not yet been integrated, although another payment capability may already exist. Depending on the project, the agent may connect an existing payment entry point or add a product-specific entry point. \textbf{Advanced (Risk-Aware Payment Hardening)} starts after the core Alipay integration is present and evaluates payment-specific safety and boundary handling. The two scenarios are therefore progressive rather than independent: Basic evaluates whether the agent can establish the end-to-end payment loop, whereas Advanced evaluates whether it can harden that loop against risks such as invalid or duplicated notifications, non-idempotent state updates, abnormal transaction states, and unsafe refund behavior.

We define a task instance as a project--scenario pair. Each task instance contains agent-facing inputs and evaluator-facing materials. The agent-facing inputs consist of an initial repository and a scenario-specific natural-language instruction, and the expected output is a modified repository. The evaluator-facing materials consist of scenario-specific rubrics, deterministic checks derived from those rubrics, and supplementary LLM-assisted criteria for rubric items requiring semantic or payment-domain judgment. Pairing each of the nine product-specific projects with Basic and Advanced scenarios yields 18 task instances. This shared format supports consistent evaluation across products, scenarios, agents, and \noskill/\withskill conditions.

\begin{table}[H]
\caption{Product and project coverage of \bench. A checkmark indicates an existing payment integration in the initial repository, while a cross indicates no initial payment integration. Basic and Advanced columns report the number of rubric items in each scenario.}
\label{tab:products}
\centering
\scriptsize
\setlength{\tabcolsep}{4pt}
\resizebox{\textwidth}{!}{%
\begin{tabular}{lllccc}
\toprule
\multicolumn{1}{c}{Product} & \multicolumn{1}{c}{Project} & \multicolumn{1}{c}{Backend stack} & \multicolumn{1}{c}{Initial Payment} & \multicolumn{1}{c}{Basic} & \multicolumn{1}{c}{Advanced} \\
\midrule
In-App Payment & ez\_tickets\_app & Node.js + Express + MySQL & \multicolumn{1}{c}{$\times$} & 21 & 41 \\
Usage Based Payment & A2M Recipes & Next.js + TypeScript + Node.js & \multicolumn{1}{c}{$\times$} & 20 & 22 \\
Mobile Web Payment & eDoc-doctor-appointment-system & PHP + Apache + MySQL & \multicolumn{1}{c}{$\times$} & 13 & 21 \\
JSAPI Payment & laravel-gymie & Laravel + PHP + SQLite & \multicolumn{1}{c}{$\times$} & 26 & 24 \\
Authorization Hold & bookcars & Node.js + TypeScript + MongoDB & \multicolumn{1}{c}{$\checkmark$} & 18 & 33 \\
PC Web Payment & litemall & Java + Spring Boot + MySQL & \multicolumn{1}{c}{$\checkmark$} & 18 & 36 \\
Order QR Code Payment & bill-express & Node.js + Express + SQLite & \multicolumn{1}{c}{$\times$} & 17 & 40 \\
QR Code Payment & bill-express & Node.js + Express + SQLite & \multicolumn{1}{c}{$\times$} & 17 & 41 \\
\mbox{Merchant Initiated Transaction} & saas-starter & Next.js + TypeScript + Postgres & \multicolumn{1}{c}{$\checkmark$} & 18 & 33 \\
\bottomrule
\end{tabular}
}
\end{table}

Table~\ref{tab:products} summarizes the product-level coverage of the current release. The project column identifies the corresponding repositories or project sites~\citep{repo_ez_tickets_app,repo_ai_collection_pilot,repo_hi_events,repo_laravel_gymie,repo_bookcars,repo_litemall,repo_ospos,repo_unibee}. For each product, the table records the project, backend technology stack, initial payment status, and the number of rubric items in the Basic and Advanced scenarios. Although product details differ, Basic tasks typically require coordinated changes to payment configuration, backend payment creation and confirmation, local business state, and application-facing payment entry points where applicable. Advanced tasks build on this foundation and require the agent to address payment-specific risks such as duplicate notifications, repeated confirmation calls, inconsistent local states, unsafe refund behavior, and insufficient verification of asynchronous notifications.

\subsection{Product-Grounded Task Construction}

\paragraph{Project and scenario initialization}
We construct task instances from business-style applications and payment-specific integration requirements. For each Alipay payment product, we select a repository in which the surrounding business workflow is already implemented. The repository's existing payment state is product-dependent: some projects contain no payment capability, whereas others contain an existing payment integration. These differences are retained as part of the task context and are summarized in Table~\ref{tab:products}.

For each project, we prepare two independently initialized scenario states. The Basic state is a repository snapshot taken before the target Alipay product is integrated. The Advanced state is a separately prepared snapshot in which the core target-product payment path is already present, so that the task can focus on product-specific safety, fund-risk, and boundary requirements. Basic and Advanced are evaluated as separate controlled task instances with their own scenario-specific instructions, rubrics, and starting repositories.

\paragraph{Instruction and rubrics}
For each product--scenario pair, we prepare a payment-specific natural-language instruction and a corresponding set of rubrics. The instruction defines the task presented to the agent. It specifies the business objective, repository context, payment behavior, and constraints that the solution must satisfy. Written as a developer-facing integration request, it generally leaves internal code organization open while specifying externally observable interfaces, payment semantics, security constraints, and, where necessary, required API, SDK, or configuration behavior. The target payment product is explicitly specified through the task metadata and corresponding instruction, which may additionally provide product-specific API and business-context details. Where relevant, the instruction also identifies the interfaces and observable behaviors used to evaluate the completed integration.

The instruction reflects the starting state of each scenario. A Basic instruction asks the agent to add or complete the target payment capability within an existing business workflow. An Advanced instruction addresses safety-sensitive or boundary requirements in a separately prepared payment-integration context. The task description and interface requirements remain specific to each product and project, with product-specific API and integration requirements grounded in the Alipay Open Platform documentation \citep{alipayopenplatform}.

The rubrics decompose each instruction into checkable requirements. Their composition varies across products and projects. Basic rubrics may assess payment creation, signing and credential handling, application-facing invocation, payment-result processing, and consistency between payment and business states. Advanced rubrics may assess safety-sensitive properties such as idempotent state transitions, duplicate-notification handling, abnormal transaction states, notification authenticity, or refund boundaries, where relevant to the task.

\begin{figure*}[t]
    \centering
    \includegraphics[width=\textwidth]{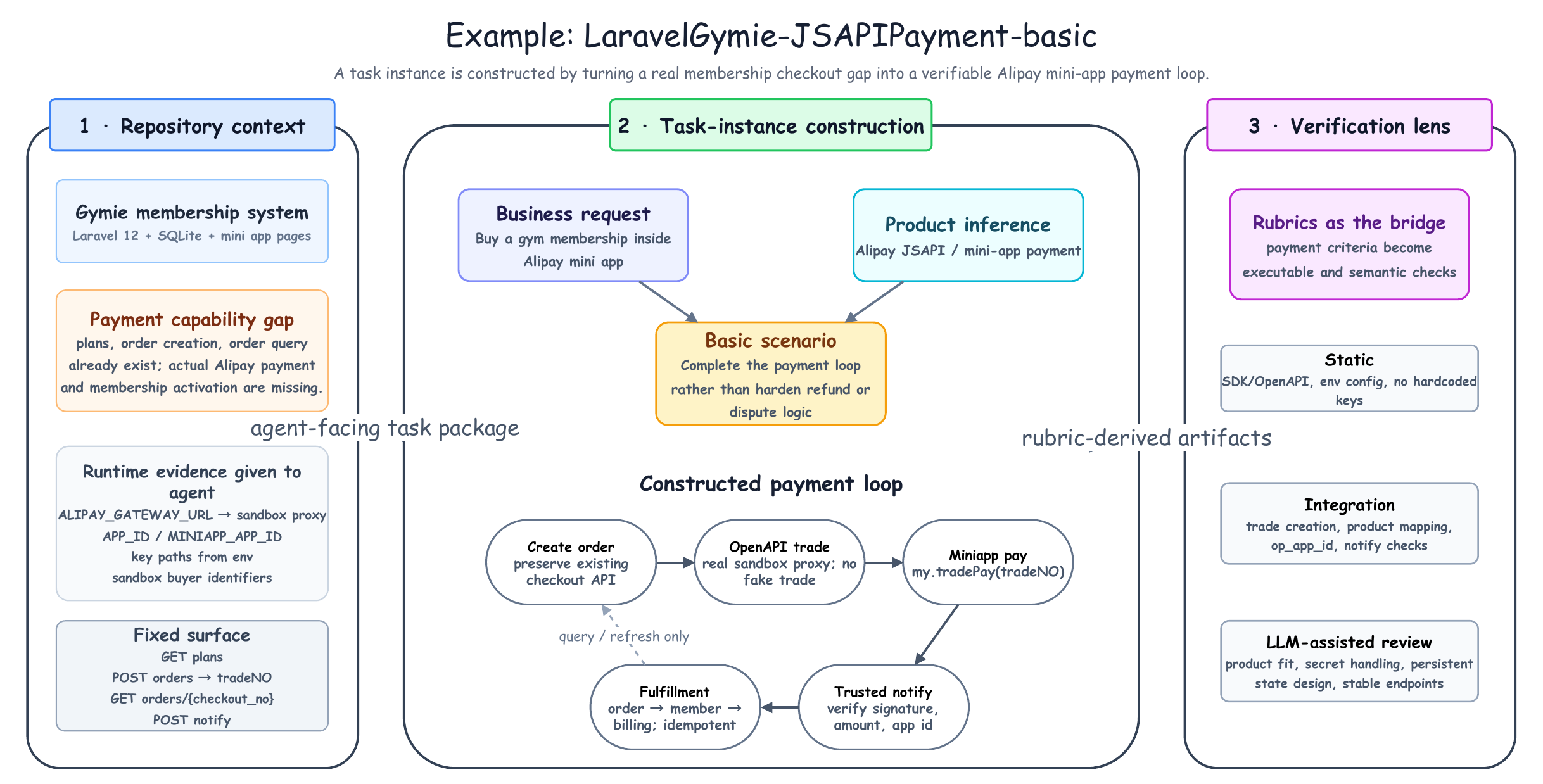}
    \caption{Running example of task construction and rubric-derived evaluation for the \texttt{LaravelGymie-JSAPIPayment-basic} task. The task covers server-side trade creation, Mini Program payment invocation, trusted notification processing, and idempotent membership fulfillment, and is evaluated under the paired \noskill/\withskill condition.}
    \label{fig:task-construction-example}
\end{figure*}

\paragraph{Running example}
Figure~\ref{fig:task-construction-example} illustrates the construction and evaluation of the \texttt{LaravelGymie-JSAPIPayment-basic} task. The initial Laravel repository provides a gym-membership workflow, including membership plans, order creation and query APIs, and an Alipay Mini Program page, but stops after creating a pending order. The instruction asks the agent to complete the existing workflow with Alipay Mini Program payment rather than introduce a replacement API. A valid solution must create the payment trade on the server, return the case-sensitive \texttt{tradeNO} field for native Mini Program invocation, and advance the order and membership fulfillment only after a trusted server-side result. The task rubrics further require invalid notifications to be rejected, repeated success notifications to be handled idempotently, and client-side callbacks not to determine the final payment state.

The same scenario-specific rubrics are instantiated as task-specific deterministic checks, including static, unit, integration, and E2E tests where applicable, with product-specific client checks where applicable, together with supplementary LLM-assisted criteria for properties that require semantic code inspection. The resulting evaluators are then reused for the paired \noskill/\withskill comparison.

\paragraph{Rubric-Derived Evaluation}
The scenario-specific rubrics serve as the source specification for evaluation. For rubric items that can be operationalized as executable checks, we derive deterministic evaluators, including static checks, unit tests where applicable, integration tests, and E2E or environment-level validation. Static checks examine structural properties such as the presence and placement of payment-related code, configuration, and credential handling. Unit and integration tests examine local payment logic, backend API behavior, state transitions, and interactions between payment and business modules.

When evaluation requires Alipay SDK calls, we use the Alipay sandbox environment where reliable support is available. When a product is not supported by the sandbox, or when rare abnormal states and risk boundaries cannot be triggered reproducibly, we use deterministic mock responses or fixtures. These cases are derived from the same scenario-specific rubrics and manually reviewed to ensure that they evaluate payment behavior rather than incidental implementation details.

For rubric items that cannot be fully captured by executable checks, we derive supplementary LLM-assisted assessment criteria from the same rubrics. These criteria examine properties such as payment-product fit, end-to-end flow consistency, security-sensitive logic placement, and payment--business state consistency. We manually review both deterministic checks and LLM-assisted criteria for consistency with the task instruction and the intended rubric item. This procedure makes the resulting evidence traceable to the same task specification rather than defining executable and semantic evaluators independently.

\paragraph{Paired skill intervention}
Finally, we define the paired intervention used to study the practical value of structured payment guidance. In the \noskill condition, the agent receives the project and task instruction without access to \payskill. In the \withskill condition, the same agent also receives the official \payskill skill, which provides reusable integration guidance rather than task-specific solutions. Its scope includes payment-product guidance, integration procedures, interface explanations, SDK and sandbox guidance, code examples, and common payment-integration pitfalls.

For each pair, we keep the agent, project, task instruction, and execution environment fixed, and evaluate both outputs using the same scenario-specific rubrics and evaluators. The paired protocol supports analysis of how access to structured Alipay payment guidance affects payment-product selection, integration efficiency, payment safety, and payment--business state consistency.

\subsection{Evaluation and Scoring}
\label{sec:evaluation-scoring}

\subsubsection{Evaluation Signals}

We evaluate each generated solution using test items aligned with its scenario-specific rubrics. The evaluator executes these checks on the modified repository. Each item is assigned to static analysis, unit testing where applicable, integration testing, E2E validation, or LLM-assisted review. 

\paragraph{Static checks.}
Static checks examine source-level and structural signals, including SDK or OpenAPI usage, credential handling, payment entry points, signature-verification capability, field binding, and state or refund models where applicable. For safety tasks, they also inspect notification-verification hooks, fake-success bypasses, idempotency or terminal-state guards, refund or query capability, and secret-management safeguards. These checks provide lightweight evidence of the expected implementation structure, but do not establish end-to-end runtime correctness.

\paragraph{Unit tests.}
Unit tests are used where a task provides applicable local test coverage, such as project-specific application tests.

\paragraph{Integration tests.}
Integration tests verify application startup, product-specific payment flows, backend state transitions, and interactions with payment gateways or deterministic fixtures. Safety tasks additionally cover duplicated notifications, repeated requests, invalid signatures, wrong amounts, exceptional transaction states, refund boundaries, and related risks where applicable. Some tasks use local mock Alipay gateways or deterministic fixtures for reproducibility rather than relying uniformly on live sandbox behavior.

\paragraph{E2E tests.}
End-to-end (E2E) tests exercise the user-facing or integration-facing entry points required by each task. They verify endpoint reachability, backend invocation, and observable states such as success, failure, cancellation, and pending where applicable. Some combined business-and-payment logic is also exercised by integration tests.

\paragraph{LLM-assisted domain assessment.}
LLM-assisted domain assessment examines rubric-defined semantic properties, including product mapping where applicable, server-side signing and confirmation, state-machine coherence, security-sensitive logic placement, and cross-component linkage. For safety tasks, it further assesses whether the implementation addresses the requested risk class rather than merely preserving the happy path. This signal helps identify plausible but unsafe implementations that deterministic checks may miss.

\subsubsection{Rubric-Level Metric and Aggregation}
\label{sec:rpr-aggregation}

We use Rubric Pass Rate (RPR) as the primary metric. RPR is the fraction of rubric requirements satisfied by a generated solution. A configuration $c$ denotes a model evaluated under one skill condition, either \noskill or \withskill. For task instance $t$ and trial $r$, let $\mathcal{M}_t$ denote the evaluation methods used for $t$. Let $\mathcal{I}_{t,m}$ denote the rubric items evaluated by method $m$. We encode each outcome as $y_{c,t,r,i}\in\{0,1\}$: pass is 1, and fail or evaluator error is 0. The weighted RPR is
\[
\mathrm{RPR}_{c,t,r} =
\frac{
\sum_{m \in \mathcal{M}_t}
\alpha_m
\left(
\frac{1}{|\mathcal{I}_{t,m}|}
\sum_{i \in \mathcal{I}_{t,m}}
y_{c,t,r,i}
\right)}
{\sum_{m \in \mathcal{M}_t} \alpha_m},
\]
where $\alpha_m$ is the relative weight of method $m$. We use conservative relative weights:
\[
\alpha_{\mathrm{static}}=\alpha_{\mathrm{unit}}=\alpha_{\mathrm{llm}}=1,
\qquad
\alpha_{\mathrm{integration}}=\alpha_{\mathrm{e2e}}=2.
\]
The denominator normalizes over the methods available for $t$. Thus, RPR remains in $[0,1]$ when task instances use different evaluator sets. We assign higher weight to integration and end-to-end checks because they more directly test whether the generated repository preserves a coherent payment flow across backend services, application state and user-facing entry points. Static, unit and LLM-assisted checks provide complementary but more localized evidence.

Reported scores are macro-averages over project-scenario task instances. For repeated trials of the same task instance, we first compute its mean RPR, so each task instance contributes equally regardless of its rubric count. LLM-assisted domain assessment enters the same aggregation as a rubric-aligned method. Evaluator errors are also retained separately for diagnostic analysis.

\section{Experiments and Results}

To assess the performance of coding agents on realistic Alipay payment integration tasks, we evaluated six models on all 18 task instances in \bench. Each instance paired a project with a scenario and was evaluated under paired conditions with and without access to \payskill. The experiments were organized around three research questions:
\begin{itemize}
    \item \textbf{RQ1: Model capability.} How capable are current models at completing Alipay payment integration across products and scenarios?
    \item \textbf{RQ2: Skill intervention.} How does access to \payskill change performance?
    \item \textbf{RQ3: Evaluation-method diagnostics.} What aspects of Alipay payment integration capability are revealed by different evaluation methods?
\end{itemize}

\subsection{Experimental Setup}

We conducted all experiments on \bench using Claude Code version 2.1.200. The evaluation covered six high-performing model variants available in our environment at the time of the study: \claudeopus, \glm, \kimicode, \deepseekpro, \minimax, and \qwenthirlyseven. Official documentation is available for the Claude Opus 4.8, GLM-5.2, Kimi K2.7 Code, DeepSeek V4, and MiniMax M3 variants used in this study \citep{anthropic2026opus48,zai2026glm52,moonshot2026kimi27code,deepseek2026v4preview,minimax2026m3}.
Each trial started from a fresh Docker container initialized with the repository and evaluation environment for the corresponding task. Consequently, files, dependency caches, service states, and evaluator outputs generated during one trial did not affect any other trial. For each combination of model, project, scenario, and skill condition, we kept the task configuration and evaluation procedure fixed.

\subsection{RQ1: Model Capability}

\begin{figure}[t]
\centering
\includegraphics[width=\textwidth]{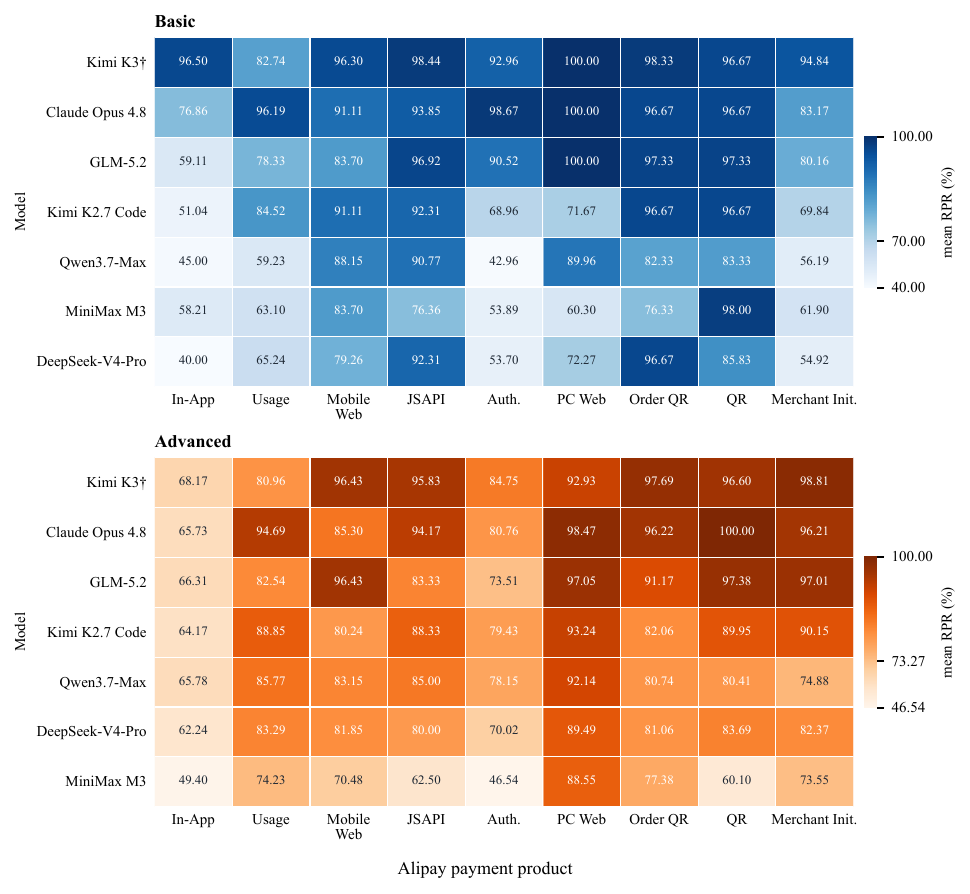}
\caption{Model capability across Alipay payment integration products and scenarios under the \withskill condition. Each cell reports mean RPR for one model, product, and scenario. The top and bottom panels show Basic and Advanced tasks, respectively, using separate color scales.}
\label{fig:rq1-withskill-heatmap}
\end{figure}

RQ1 characterizes model capability across Alipay payment integration products and scenarios. The results reported here come from configurations in which \payskill was installed in the coding agent. Under this setting, the comparison focuses on how models use repository context, payment-domain instructions, and project-specific code structure. Figure~\ref{fig:rq1-withskill-heatmap} reports mean RPR for each model, product, and scenario. The top and bottom panels correspond to Basic and Advanced tasks, respectively, with an independent color scale in each panel.

First, model capability varies substantially across Alipay payment integration tasks. \claudeopus achieves the highest overall mean RPR (91.37\%), followed by \glm (87.12\%) and \kimicode (82.18\%). \qwenthirlyseven and \deepseekpro obtain lower intermediate scores of 75.78\% and 75.23\%, respectively, while \minimax achieves 68.58\%. The 22.79 percentage-point gap between the highest and lowest scores indicates substantial variation in model capability.

Second, this variation is also reflected in performance consistency across products and scenarios. \claudeopus achieves at least 90\% mean RPR in 13 of the 18 product--scenario cells, and \glm reaches this level in 10 cells. \kimicode reaches the same level in 6 cells, while \qwenthirlyseven and \deepseekpro do so in 2 cells each. At the same time, even \claudeopus falls to 65.73\% on Advanced \appzhifu, whereas \minimax ranges from 46.54\% on Advanced \yushouquanzhifu to 98.00\% on Basic \dangmianfu. These results show that model capability is expressed through both overall performance and consistency across payment products and scenarios.

Third, product context shapes how model capability is expressed. App has the lowest overall score (58.65\%), and only \claudeopus exceeds 65\% mean RPR on this product (71.30\%). The source application originally supported online booking followed by offline settlement, rather than an existing online payment flow. Agents therefore had to introduce a payment entry point, backend payment state, and confirmation logic into a workflow not organized around online settlement. The other models score lower on App, including \glm at 62.71\%, \qwenthirlyseven at 55.39\%, and \deepseekpro at 51.12\%. Pre-Authorization exhibits a different profile. Unlike In-App Payment, which is challenging for nearly all models, Pre-Authorization shows substantial variation across models and scenarios. Its overall mean RPR is 69.76\%, while individual scores range from 42.96\% to 98.67\%. These results show that some products are difficult for nearly every model, whereas others differentiate models by their ability to maintain payment state, confirmation, and risk-handling logic across a longer code path.

Finally, Basic and Advanced tasks provide complementary views of model capability. Basic tasks primarily test whether a model can establish the initial Alipay payment loop. Advanced tasks focus on fund safety, risk handling, and boundary conditions after a basic integration exists. The scenario split therefore reveals different model profiles rather than a single difficulty ordering. \claudeopus remains strong in both scenarios, with scores of 92.58\% on Basic tasks and 90.17\% on Advanced tasks. \glm is similarly stable, scoring 87.05\% and 87.19\%, respectively. By contrast, \qwenthirlyseven scores lower on Basic tasks (70.88\%) than on Advanced tasks (80.67\%), whereas \minimax performs better on Basic tasks than on Advanced tasks (70.20\% versus 66.97\%). These results indicate that constructing a payment flow and preserving safety-related payment behavior should be examined as distinct capabilities.

\noindent\textbf{RQ1 Summary. Model capability explains much of the observed performance spread and is expressed through overall performance, cross-product consistency, and adaptation to scenario-specific requirements. \bench therefore supports a fine-grained characterization of current models, showing where their Alipay payment integration capability is strong and where it remains limited.}

\subsection{RQ2: Skill Intervention}

RQ2 examines how access to \payskill changes coding-agent performance on Alipay payment integration tasks. We compare paired \noskill and \withskill trials under the same task, repository, and evaluation procedure. Table~\ref{tab:rq2-skill-effect-performance} reports mean RPR for each model, product, and scenario, together with the percentage-point change associated with access to \payskill.

\begin{table}[!t]
\caption{Main results for skill intervention: mean RPR (\%) with and without \payskill across models, products, and scenarios. The task, repository, and evaluator are fixed within each paired setting; only access to \payskill is varied. For each scenario--model block, No and With report the \noskill and \withskill conditions, and $\Delta$ reports the percentage-point change, defined as $100 \times$ (\withskill mean RPR minus \noskill mean RPR). Positive and negative changes are shaded in blue and orange.}
\label{tab:rq2-skill-effect-performance}
\begin{center}
\begingroup
\tiny
\renewcommand{\arraystretch}{0.96}
\setlength{\tabcolsep}{2.4pt}
\setlength{\aboverulesep}{0.25ex}
\setlength{\belowrulesep}{0.25ex}
\resizebox{\textwidth}{!}{%
\begin{tabular}{clccccccccccc}
\toprule
\rqhead{Scenario} & \rqhead{Model} & \rqhead{Skill\\Intervention} & \rqhead{In-App} & \rqhead{Usage} & \rqhead{Mobile\\Web} & \rqhead{JSAPI} & \rqhead{Auth.} & \rqhead{PC\\Web} & \rqhead{Order QR} & \rqhead{QR} & \rqhead{Merchant\\Init.} & \rqhead{Avg.} \\
\midrule
\multirow{18}{*}{\textbf{Basic}} & \multirow{3}{*}{\claudeopus} & without & 50.3 & 92.4 & 91.1 & 80.8 & 89.2 & 86.7 & 96.7 & 96.7 & 71.7 & 84.0 \\
 &  & with & 76.9 & 96.2 & 91.1 & 93.8 & 98.7 & 100.0 & 96.7 & 96.7 & 83.2 & 92.6 \\
 &  & $\Delta$ & \deltaposhigh{+26.5} & \deltaposlow{+3.8} & 0.0 & \deltaposmid{+13.0} & \deltaposmid{+9.4} & \deltaposmid{+13.3} & 0.0 & 0.0 & \deltaposmid{+11.4} & \deltaavg{\deltaposmid{+8.6}} \\
\cmidrule[0.2pt](lr){3-13}
 & \multirow{3}{*}{\glm} & without & 50.6 & 66.2 & 79.3 & 91.1 & 59.3 & 90.9 & 96.7 & 90.8 & 65.7 & 76.7 \\
 &  & with & 59.1 & 78.3 & 83.7 & 96.9 & 90.5 & 100.0 & 97.3 & 97.3 & 80.2 & 87.0 \\
 &  & $\Delta$ & \deltaposmid{+8.5} & \deltaposmid{+12.1} & \deltaposlow{+4.4} & \deltaposmid{+5.8} & \deltaposhigh{+31.3} & \deltaposmid{+9.1} & \deltaposlow{+0.7} & \deltaposmid{+6.5} & \deltaposmid{+14.4} & \deltaavg{\deltaposmid{+10.3}} \\
\cmidrule[0.2pt](lr){3-13}
 & \multirow{3}{*}{\kimicode} & without & 44.3 & 72.4 & 74.8 & 65.4 & 60.3 & 60.6 & 75.0 & 90.0 & 55.9 & 66.5 \\
 &  & with & 51.0 & 84.5 & 91.1 & 92.3 & 69.0 & 71.7 & 96.7 & 96.7 & 69.8 & 80.3 \\
 &  & $\Delta$ & \deltaposmid{+6.7} & \deltaposmid{+12.1} & \deltaposhigh{+16.3} & \deltaposhigh{+26.9} & \deltaposmid{+8.6} & \deltaposmid{+11.1} & \deltaposhigh{+21.7} & \deltaposmid{+6.7} & \deltaposmid{+14.0} & \deltaavg{\deltaposmid{+13.8}} \\
\cmidrule[0.2pt](lr){3-13}
 & \multirow{3}{*}{\deepseekpro} & without & 35.7 & 42.9 & 71.9 & 83.1 & 37.1 & 64.2 & 58.3 & 60.8 & 45.1 & 55.5 \\
 &  & with & 40.0 & 65.2 & 79.3 & 92.3 & 53.7 & 72.3 & 96.7 & 85.8 & 54.9 & 71.1 \\
 &  & $\Delta$ & \deltaposlow{+4.3} & \deltaposhigh{+22.4} & \deltaposmid{+7.4} & \deltaposmid{+9.2} & \deltaposhigh{+16.6} & \deltaposmid{+8.0} & \deltaposhigh{+38.3} & \deltaposhigh{+25.0} & \deltaposmid{+9.8} & \deltaavg{\deltaposhigh{+15.7}} \\
\cmidrule[0.2pt](lr){3-13}
 & \multirow{3}{*}{\minimax} & without & 43.2 & 60.7 & 73.3 & 71.8 & 47.9 & 59.5 & 52.3 & 96.5 & 48.3 & 61.5 \\
 &  & with & 58.2 & 63.1 & 83.7 & 76.4 & 53.9 & 60.3 & 76.3 & 98.0 & 61.9 & 70.2 \\
 &  & $\Delta$ & \deltaposmid{+15.0} & \deltaposlow{+2.4} & \deltaposmid{+10.4} & \deltaposlow{+4.5} & \deltaposmid{+6.0} & \deltaposlow{+0.8} & \deltaposhigh{+24.0} & \deltaposlow{+1.5} & \deltaposmid{+13.7} & \deltaavg{\deltaposmid{+8.7}} \\
\cmidrule[0.2pt](lr){3-13}
 & \multirow{3}{*}{\qwenthirlyseven} & without & 33.7 & 50.9 & 73.3 & 62.7 & 32.9 & 81.2 & 75.8 & 81.7 & 51.1 & 60.4 \\
 &  & with & 45.0 & 59.2 & 88.1 & 90.8 & 43.0 & 90.0 & 82.3 & 83.3 & 56.2 & 70.9 \\
 &  & $\Delta$ & \deltaposmid{+11.3} & \deltaposmid{+8.3} & \deltaposmid{+14.8} & \deltaposhigh{+28.1} & \deltaposmid{+10.0} & \deltaposmid{+8.7} & \deltaposmid{+6.5} & \deltaposlow{+1.7} & \deltaposmid{+5.1} & \deltaavg{\deltaposmid{+10.5}} \\
\cmidrule(lr){1-13}
\multirow{18}{*}{\textbf{Advanced}} & \multirow{3}{*}{\claudeopus} & without & 60.3 & 84.1 & 82.4 & 93.3 & 71.6 & 97.9 & 90.7 & 96.4 & 89.9 & 85.2 \\
 &  & with & 65.7 & 94.7 & 85.3 & 94.2 & 80.8 & 98.5 & 96.2 & 100.0 & 96.2 & 90.2 \\
 &  & $\Delta$ & \deltaposmid{+5.4} & \deltaposmid{+10.6} & \deltaposlow{+2.9} & \deltaposlow{+0.8} & \deltaposmid{+9.2} & \deltaposlow{+0.6} & \deltaposmid{+5.6} & \deltaposlow{+3.6} & \deltaposmid{+6.3} & \deltaavg{\deltaposlow{+5.0}} \\
\cmidrule[0.2pt](lr){3-13}
 & \multirow{3}{*}{\glm} & without & 63.8 & 71.0 & 90.0 & 72.5 & 69.9 & 91.3 & 78.0 & 93.3 & 93.7 & 80.4 \\
 &  & with & 66.3 & 82.5 & 96.4 & 83.3 & 73.5 & 97.0 & 91.2 & 97.4 & 97.0 & 87.2 \\
 &  & $\Delta$ & \deltaposlow{+2.5} & \deltaposmid{+11.6} & \deltaposmid{+6.4} & \deltaposmid{+10.8} & \deltaposlow{+3.6} & \deltaposmid{+5.7} & \deltaposmid{+13.2} & \deltaposlow{+4.1} & \deltaposlow{+3.3} & \deltaavg{\deltaposmid{+6.8}} \\
\cmidrule[0.2pt](lr){3-13}
 & \multirow{3}{*}{\kimicode} & without & 45.4 & 73.5 & 83.9 & 65.8 & 31.4 & 91.7 & 63.5 & 69.7 & 76.4 & 66.8 \\
 &  & with & 64.2 & 88.8 & 80.2 & 88.3 & 79.4 & 93.2 & 82.1 & 89.9 & 90.2 & 84.0 \\
 &  & $\Delta$ & \deltaposhigh{+18.7} & \deltaposhigh{+15.4} & \deltaneglow{-3.7} & \deltaposhigh{+22.5} & \deltaposhigh{+48.0} & \deltaposlow{+1.6} & \deltaposhigh{+18.5} & \deltaposhigh{+20.3} & \deltaposmid{+13.8} & \deltaavg{\deltaposhigh{+17.2}} \\
\cmidrule[0.2pt](lr){3-13}
 & \multirow{3}{*}{\deepseekpro} & without & 51.9 & 73.2 & 76.4 & 75.0 & 63.7 & 91.3 & 77.2 & 75.5 & 57.4 & 71.3 \\
 &  & with & 62.2 & 83.3 & 81.8 & 80.0 & 70.0 & 89.5 & 81.1 & 83.7 & 82.4 & 79.3 \\
 &  & $\Delta$ & \deltaposmid{+10.3} & \deltaposmid{+10.1} & \deltaposmid{+5.5} & \deltaposlow{+5.0} & \deltaposmid{+6.3} & \deltaneglow{-1.8} & \deltaposlow{+3.8} & \deltaposmid{+8.2} & \deltaposhigh{+25.0} & \deltaavg{\deltaposmid{+8.0}} \\
\cmidrule[0.2pt](lr){3-13}
 & \multirow{3}{*}{\minimax} & without & 33.7 & 67.5 & 72.4 & 55.8 & 51.8 & 78.7 & 67.0 & 45.4 & 64.9 & 59.7 \\
 &  & with & 49.4 & 74.2 & 70.5 & 62.5 & 46.5 & 88.5 & 77.4 & 60.1 & 73.5 & 67.0 \\
 &  & $\Delta$ & \deltaposhigh{+15.7} & \deltaposmid{+6.8} & \deltaneglow{-2.0} & \deltaposmid{+6.7} & \deltanegmid{-5.3} & \deltaposmid{+9.8} & \deltaposmid{+10.4} & \deltaposmid{+14.7} & \deltaposmid{+8.6} & \deltaavg{\deltaposmid{+7.3}} \\
\cmidrule[0.2pt](lr){3-13}
 & \multirow{3}{*}{\qwenthirlyseven} & without & 52.3 & 76.2 & 66.6 & 75.0 & 58.8 & 65.7 & 75.5 & 78.9 & 70.8 & 68.9 \\
 &  & with & 65.8 & 85.8 & 83.2 & 85.0 & 78.2 & 92.1 & 80.7 & 80.4 & 74.9 & 80.7 \\
 &  & $\Delta$ & \deltaposmid{+13.5} & \deltaposmid{+9.6} & \deltaposhigh{+16.5} & \deltaposmid{+10.0} & \deltaposhigh{+19.4} & \deltaposhigh{+26.4} & \deltaposmid{+5.2} & \deltaposlow{+1.5} & \deltaposlow{+4.0} & \deltaavg{\deltaposmid{+11.8}} \\
\bottomrule
\end{tabular}%
}
\endgroup
\end{center}
\end{table}

\newcommand{\tokenvalidationtable}{%
\begin{table}[!t]
\caption{Output-normalized effectiveness under skill intervention. No and With report output tokens per unit RPR, computed as output tokens in K divided by mean RPR expressed on a 0--1 scale, for the \noskill and \withskill conditions, respectively. The relative row reports the paired ratio $(O/\mathrm{RPR})_{\withskill}/(O/\mathrm{RPR})_{\noskill}$ for the same model--product--scenario cell. Values below 1 indicate fewer generated output tokens per unit RPR under \withskill. Blue and orange shading mark lower and higher relative output/RPR, respectively.}
\label{tab:rq2-output-normalized}
\begin{center}
\begingroup
\tiny
\renewcommand{\arraystretch}{0.96}
\setlength{\tabcolsep}{2.4pt}
\setlength{\aboverulesep}{0.25ex}
\setlength{\belowrulesep}{0.25ex}
\resizebox{\textwidth}{!}{%
\begin{tabular}{clccccccccccc}
\toprule
\rqhead{Scenario} & \rqhead{Model} & \rqhead{} & \rqhead{In-App} & \rqhead{Usage} & \rqhead{Mobile\\Web} & \rqhead{JSAPI} & \rqhead{Auth.} & \rqhead{PC\\Web} & \rqhead{Order QR} & \rqhead{QR} & \rqhead{Merchant\\Init.} & \rqhead{Avg.} \\
\midrule
\multirow{18}{*}{\textbf{Basic}} & \multirow{3}{*}{\claudeopus} & without & 133.1 & 31.2 & 40.0 & 79.6 & 82.0 & 40.1 & 46.0 & 55.9 & 73.4 & 64.6 \\
 &  & with & 95.5 & 36.2 & 35.8 & 76.9 & 79.6 & 25.6 & 51.4 & 53.7 & 87.6 & 60.3 \\
 &  & relative & \outputrelhigh{0.72} & \outputbadmid{1.16} & \outputrelmid{0.90} & \outputrellow{0.97} & \outputrellow{0.97} & \outputrelhigh{0.64} & \outputbadlow{1.12} & \outputrellow{0.96} & \outputbadmid{1.19} & \textbf{\outputrellow{0.93}} \\
\cmidrule[0.2pt](lr){3-13}
 & \multirow{3}{*}{\glm} & without & 85.1 & 21.8 & 36.6 & 56.0 & 68.7 & 22.2 & 39.6 & 37.5 & 42.4 & 45.5 \\
 &  & with & 75.0 & 24.5 & 49.2 & 43.4 & 56.8 & 18.8 & 33.4 & 41.4 & 43.0 & 42.8 \\
 &  & relative & \outputrelmid{0.88} & \outputbadlow{1.12} & \outputbadmid{1.34} & \outputrelmid{0.78} & \outputrelmid{0.83} & \outputrelmid{0.85} & \outputrelmid{0.85} & \outputbadlow{1.10} & \outputbadlow{1.01} & \textbf{\outputrellow{0.94}} \\
\cmidrule[0.2pt](lr){3-13}
 & \multirow{3}{*}{\kimicode} & without & 84.8 & 18.7 & 36.3 & 76.1 & 58.6 & 34.8 & 35.2 & 37.9 & 70.5 & 50.3 \\
 &  & with & 66.3 & 21.5 & 36.9 & 52.9 & 34.8 & 23.4 & 33.8 & 30.8 & 57.5 & 39.8 \\
 &  & relative & \outputrelmid{0.78} & \outputbadlow{1.15} & \outputbadlow{1.02} & \outputrelhigh{0.70} & \outputrelhigh{0.59} & \outputrelhigh{0.67} & \outputrellow{0.96} & \outputrelmid{0.81} & \outputrelmid{0.82} & \textbf{\outputrelmid{0.79}} \\
\cmidrule[0.2pt](lr){3-13}
 & \multirow{3}{*}{\deepseekpro} & without & 87.2 & 28.9 & 35.1 & 61.1 & 85.9 & 24.8 & 28.2 & 46.1 & 52.1 & 49.9 \\
 &  & with & 76.9 & 19.9 & 27.7 & 30.0 & 59.1 & 23.1 & 19.7 & 22.4 & 41.1 & 35.6 \\
 &  & relative & \outputrelmid{0.88} & \outputrelhigh{0.69} & \outputrelmid{0.79} & \outputrelhigh{0.49} & \outputrelhigh{0.69} & \outputrellow{0.93} & \outputrelhigh{0.70} & \outputrelhigh{0.49} & \outputrelmid{0.79} & \textbf{\outputrelhigh{0.71}} \\
\cmidrule[0.2pt](lr){3-13}
 & \multirow{3}{*}{\minimax} & without & 88.1 & 42.3 & 47.9 & 90.5 & 68.8 & 53.0 & 74.4 & 52.3 & 72.6 & 65.5 \\
 &  & with & 58.7 & 48.3 & 49.2 & 57.9 & 76.9 & 32.9 & 46.5 & 37.5 & 66.6 & 52.7 \\
 &  & relative & \outputrelhigh{0.67} & \outputbadlow{1.14} & \outputbadlow{1.03} & \outputrelhigh{0.64} & \outputbadlow{1.12} & \outputrelhigh{0.62} & \outputrelhigh{0.63} & \outputrelhigh{0.72} & \outputrellow{0.92} & \textbf{\outputrelmid{0.80}} \\
\cmidrule[0.2pt](lr){3-13}
 & \multirow{3}{*}{\qwenthirlyseven} & without & 98.1 & 28.6 & 33.7 & 60.2 & 105.3 & 23.9 & 34.9 & 39.7 & 72.3 & 55.2 \\
 &  & with & 64.3 & 27.4 & 25.9 & 43.8 & 81.7 & 25.2 & 26.5 & 38.3 & 45.8 & 42.1 \\
 &  & relative & \outputrelhigh{0.66} & \outputrellow{0.96} & \outputrelmid{0.77} & \outputrelhigh{0.73} & \outputrelmid{0.78} & \outputbadlow{1.06} & \outputrelmid{0.76} & \outputrellow{0.96} & \outputrelhigh{0.63} & \textbf{\outputrelmid{0.76}} \\
\cmidrule(lr){1-13}
\multirow{18}{*}{\textbf{Advanced}} & \multirow{3}{*}{\claudeopus} & without & 134.7 & 36.4 & 40.9 & 73.0 & 64.0 & 56.9 & 85.4 & 88.0 & 116.7 & 77.3 \\
 &  & with & 116.5 & 52.0 & 46.5 & 71.4 & 67.8 & 68.4 & 89.9 & 65.2 & 89.2 & 74.1 \\
 &  & relative & \outputrelmid{0.86} & \outputbadmid{1.43} & \outputbadlow{1.14} & \outputrellow{0.98} & \outputbadlow{1.06} & \outputbadmid{1.20} & \outputbadlow{1.05} & \outputrelhigh{0.74} & \outputrelmid{0.76} & \textbf{\outputrellow{0.96}} \\
\cmidrule[0.2pt](lr){3-13}
 & \multirow{3}{*}{\glm} & without & 96.3 & 36.9 & 36.4 & 59.3 & 62.5 & 70.8 & 59.0 & 55.3 & 86.6 & 62.5 \\
 &  & with & 89.1 & 26.3 & 34.2 & 48.8 & 57.2 & 44.0 & 69.7 & 53.3 & 67.6 & 54.5 \\
 &  & relative & \outputrellow{0.93} & \outputrelhigh{0.71} & \outputrellow{0.94} & \outputrelmid{0.82} & \outputrellow{0.92} & \outputrelhigh{0.62} & \outputbadmid{1.18} & \outputrellow{0.96} & \outputrelmid{0.78} & \textbf{\outputrelmid{0.87}} \\
\cmidrule[0.2pt](lr){3-13}
 & \multirow{3}{*}{\kimicode} & without & 81.1 & 36.1 & 42.6 & 67.0 & 125.1 & 35.4 & 75.3 & 66.1 & 80.8 & 67.7 \\
 &  & with & 63.6 & 29.1 & 42.6 & 47.2 & 39.0 & 37.1 & 52.6 & 52.3 & 55.8 & 46.6 \\
 &  & relative & \outputrelmid{0.78} & \outputrelmid{0.81} & 1.00 & \outputrelhigh{0.70} & \outputrelhigh{0.31} & \outputbadlow{1.05} & \outputrelhigh{0.70} & \outputrelmid{0.79} & \outputrelhigh{0.69} & \textbf{\outputrelhigh{0.69}} \\
\cmidrule[0.2pt](lr){3-13}
 & \multirow{3}{*}{\deepseekpro} & without & 72.8 & 18.0 & 23.6 & 32.8 & 35.7 & 25.0 & 52.8 & 41.2 & 67.2 & 41.0 \\
 &  & with & 36.1 & 14.2 & 22.5 & 32.4 & 31.4 & 20.7 & 36.8 & 31.3 & 33.9 & 28.8 \\
 &  & relative & \outputrelhigh{0.50} & \outputrelmid{0.79} & \outputrellow{0.95} & \outputrellow{0.99} & \outputrelmid{0.88} & \outputrelmid{0.83} & \outputrelhigh{0.70} & \outputrelmid{0.76} & \outputrelhigh{0.50} & \textbf{\outputrelhigh{0.70}} \\
\cmidrule[0.2pt](lr){3-13}
 & \multirow{3}{*}{\minimax} & without & 128.5 & 57.8 & 51.8 & 88.8 & 54.1 & 33.6 & 82.6 & 122.6 & 118.6 & 82.0 \\
 &  & with & 77.1 & 44.5 & 54.6 & 81.0 & 94.8 & 43.7 & 47.0 & 86.6 & 55.9 & 65.0 \\
 &  & relative & \outputrelhigh{0.60} & \outputrelmid{0.77} & \outputbadlow{1.05} & \outputrellow{0.91} & \outputbadhigh{1.75} & \outputbadmid{1.30} & \outputrelhigh{0.57} & \outputrelhigh{0.71} & \outputrelhigh{0.47} & \textbf{\outputrelmid{0.79}} \\
\cmidrule[0.2pt](lr){3-13}
 & \multirow{3}{*}{\qwenthirlyseven} & without & 65.5 & 18.0 & 29.1 & 47.4 & 67.1 & 42.8 & 48.0 & 82.7 & 83.6 & 53.8 \\
 &  & with & 53.6 & 14.2 & 25.2 & 40.9 & 46.6 & 32.4 & 57.6 & 64.7 & 54.3 & 43.3 \\
 &  & relative & \outputrelmid{0.82} & \outputrelmid{0.79} & \outputrelmid{0.86} & \outputrelmid{0.86} & \outputrelhigh{0.69} & \outputrelmid{0.76} & \outputbadmid{1.20} & \outputrelmid{0.78} & \outputrelhigh{0.65} & \textbf{\outputrelmid{0.80}} \\
\bottomrule
\end{tabular}%
}
\endgroup
\end{center}
\end{table}
}

Across all 108 model--product--scenario cells, access to \payskill is associated with a 10.31 percentage-point increase in mean RPR. The paired change is positive in 101 cells, lower in 4 cells, and effectively unchanged in 3 cells. The gain is larger for Basic tasks than for Advanced tasks (+11.27 versus +9.35 points), but both scenarios benefit. Every evaluated model also improves on average, with gains ranging from +6.81 points for \claudeopus to +15.51 points for \kimicode. These results show a clear positive association under the paired evaluation protocol, with the magnitude varying by model, product, and scenario.

The gain is larger when the \noskill baseline is low. Cells with a \noskill mean RPR below 40\% improve by 17.64 points on average, compared with 4.52 points for cells with baseline scores between 80\% and 100\%. This pattern suggests that \payskill is particularly useful when the baseline solution has not recovered the payment-flow structure, although the smaller gains at high baseline scores may also reflect a ceiling effect. Representative examples include \kimicode on Advanced Pre-Authorization, which improves from 31.43\% to 79.43\% (+47.99 points), and \deepseekpro on Basic Barcode, which improves from 58.33\% to 96.67\% (+38.33 points). These examples indicate that \payskill is especially useful when models must recover product selection, payment-state transitions, payment confirmation, and callback or notification handling from repository context.
\tokenvalidationtable
The few lower-scoring cells are concentrated in Advanced tasks, where evaluation emphasizes safety and boundary behavior rather than simply adding a missing payment entry point. The remaining decreases are isolated across products and models, rather than indicating a consistent degradation under \withskill. Manual inspection of the corresponding execution traces indicates that these cases are associated with runtime interruptions or implementation failures in individual trials.

As a secondary analysis, we examined output tokens normalized by mean RPR as a
measure of generation efficiency. For each model--scenario pair, we first
averaged output tokens per unit RPR across the nine payment products and then
computed the ratio between the \withskill and \noskill conditions. All 12
model--scenario ratios were below 1.0, ranging from 0.69 to 0.96, with a mean of
0.81. This result indicates that access to \payskill consistently reduced
output tokens per unit RPR at the model--scenario aggregate level. However, some
individual product-level cells had ratios above 1.0, showing that the efficiency
gain was broad but not universal across products. Because this metric captures
only output tokens relative to rubric completion, it should be interpreted as a
relative generation-efficiency measure rather than an end-to-end estimate of
integration time or cost. 

\noindent\textbf{RQ2 Summary. Access to \payskill helps coding agents complete Alipay payment integration more
effectively and efficiently. It improves mean RPR across models and products,
with the largest gains occurring when models cannot reconstruct
product-specific payment logic from repository context alone. It also tends to
reduce output tokens per unit RPR, suggesting that structured Alipay guidance
can reduce trial and error and support a more efficient integration process for
developers using coding agents.}

\subsection{RQ3: Evaluation-Method Diagnostics}

\begin{figure}[t]
\centering
\includegraphics[width=\textwidth]{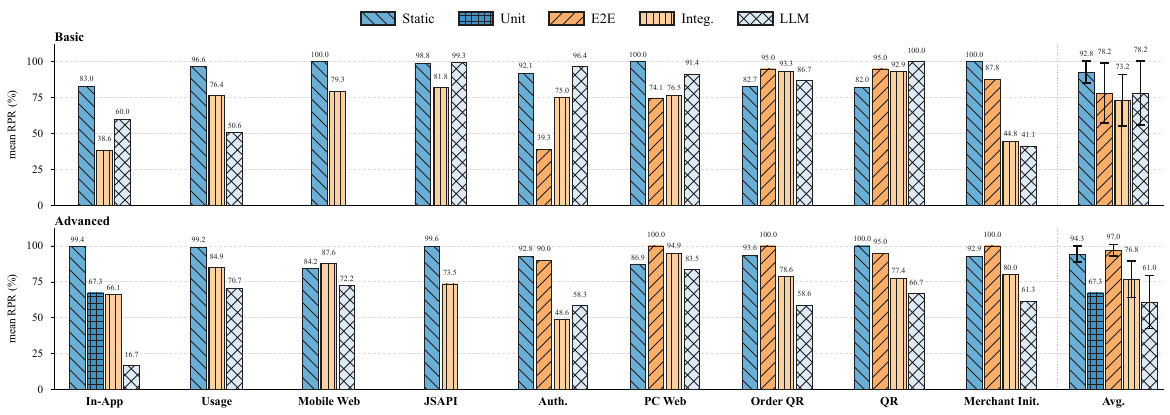}
\caption{Evaluation-method diagnostics: method-specific RPR (\%) by product and scenario under the \withskill condition. Bars report evaluator-method RPR before method weighting, averaged over models; missing evaluator methods are omitted. The Avg. group reports the across-product mean, with vertical lines indicating the standard deviation across products. The top and bottom panels show Basic and Advanced tasks, respectively.}
\label{fig:rq3-signal-decomposition}
\end{figure}

RQ3 examines what different evaluation methods reveal about Alipay payment
integration capability. The methods target different verification layers:
Static checks inspect source-level structure, Unit tests provide local
behavioral evidence where available, Integration tests exercise backend
interactions and state transitions, E2E tests inspect externally visible
workflow behavior, and LLM-assisted assessment examines payment-domain
semantics. Figure~\ref{fig:rq3-signal-decomposition} reports the unweighted
within-method RPR under the \withskill condition.

The first pattern is a gap between structural completion and executable
behavior. Static RPR averages 92.79\% for Basic tasks, whereas Integration and
E2E average 73.16\% and 78.24\%, respectively. For example, Basic In-App Payment
obtains 83.0\% Static RPR but only 38.6\% Integration RPR. These results show
that agents can introduce expected SDK usage, payment entry points, fields, and
state models without fully connecting them into an executable payment workflow.

The second pattern is a gap between executable paths and payment-domain
requirements. In Advanced tasks, E2E RPR averages 97.03\%, whereas LLM-assisted
domain assessment averages 61.05\%. Advanced Order QR, for example, reaches
100\% E2E RPR but only 58.6\% LLM-assisted RPR. This result indicates that an
implementation can execute through the tested entry points while still lacking
evidence for payment-safety and state-consistency requirements.

The reverse pattern also occurs. Basic Pre-Authorization obtains 96.4\% under
LLM-assisted assessment but only 39.3\% under E2E testing. An implementation may
therefore appear semantically consistent with product requirements while failing
through the required application-facing workflow. Together, these patterns show
that structural, executable, and domain-level evaluation expose different forms
of incomplete payment integration.

Because the methods cover different rubric subsets, their scores should not be
interpreted as direct comparisons of evaluator accuracy or strictness. Their
value lies in failure localization: the combined signals help determine whether
an integration is missing expected structure, fails during execution, or
remains incomplete in payment safety and business-state consistency.

\noindent\textbf{RQ3 Summary. Payment integration cannot be reliably assessed by checking only whether the
expected code is present or whether a tested path executes. Static,
execution-based, and LLM-assisted evaluations reveal structural completion,
executable workflow behavior, and payment-domain correctness from different
perspectives. Multi-method evaluation is therefore necessary not only to
determine whether an integration fails, but also to identify where it fails.}

\section{Conclusion}

Taken together, the results show that Alipay payment integration requires an agent to select an appropriate product, modify a repository across components, preserve payment state, and satisfy payment-specific safety requirements.

We developed \bench to measure these demands in realistic project repositories. The benchmark connects payment products with business workflows and existing codebases, and separates Basic functional completion from Advanced risk-aware hardening. Its rubric-derived framework combines deterministic checks with supplementary LLM-assisted assessment, while the paired \noskill/\withskill protocol measures the effect of structured payment guidance.

In our evaluation of 18 task instances and six models, mean RPR under the \withskill setting ranged from 68.58\% to 91.37\%. Relative to the corresponding \noskill condition, mean RPR increased by 10.31 percentage points on average, with gains in 101 of 108 model--product--scenario comparisons. The average improvement was larger for Basic tasks than for Advanced tasks (+11.27 versus +9.35 percentage points). The method-specific results also separated source-level completion from executable payment behavior and payment-domain requirements, supporting the use of progressive scenarios and complementary rubric-grounded signals.

Our results are grounded in the Alipay Open Platform products, task instances, and coding-agent configurations studied here. Future work can extend the benchmark with additional models and agent frameworks, broader repository coverage, and longer-horizon executable payment workflows, allowing the robustness of these findings to be examined across a wider range of settings.

\bibliographystyle{assets/plainnat}
\bibliography{references}

\appendix


\clearpage

\end{document}